\pdfoutput=1
\documentclass[11pt]{article}

\usepackage{ACL2023}
\usepackage{times}
\usepackage{latexsym}

\usepackage[T1]{fontenc}
\usepackage[utf8]{inputenc}

\usepackage{microtype}

\usepackage{inconsolata}

\usepackage{booktabs}
\usepackage{multirow}
\usepackage{amsmath}
\usepackage{graphicx}
\usepackage{latexsym}
\usepackage{marvosym}

\title{MedKP: Medical Dialogue with Knowledge Enhancement and Clinical Pathway Encoding}

\author{
    Jiageng Wu\textsuperscript{1}, Xian Wu\textsuperscript{2, \Letter}, Yefeng Zheng\textsuperscript{2}, Jie Yang\textsuperscript{3, \Letter} \\
    \textsuperscript{1}Zhejiang University, Hangzhou, China \\
    \textsuperscript{2}Tencent Youtu Lab, Jarvis Research Center, Shenzhen, China\\
    \textsuperscript{3}Harvard University, Boston MA, USA \\
    \texttt{jiagengwu@zju.edu.cn}, \texttt{kevinxwu@tencent.com}, \\
    \texttt{yefengzheng@tencent.com}, \texttt{jyang66@bwh.harvard.edu}
}

\begin{document}
\maketitle
\begin{abstract}
With appropriate data selection and training techniques, Large Language Models (LLMs) have demonstrated exceptional success in various medical examinations and multiple-choice questions. However, the application of LLMs in medical dialogue generation—a task more closely aligned with actual medical practice—has been less explored. This gap is attributed to the insufficient medical knowledge of LLMs, which leads to inaccuracies and hallucinated information in the generated medical responses. In this work, we introduce the Medical dialogue with Knowledge enhancement and clinical Pathway encoding (MedKP) framework, which integrates an external knowledge enhancement module through a medical knowledge graph and an internal clinical pathway encoding via medical entities and physician actions. Evaluated with comprehensive metrics, our experiments on two large-scale, real-world online medical consultation datasets (MedDG and KaMed) demonstrate that MedKP surpasses multiple baselines and mitigates the incidence of hallucinations, achieving a new state-of-the-art. Extensive ablation studies further reveal the effectiveness of each component of MedKP. This enhancement advances the development of reliable, automated medical consultation responses using LLMs, thereby broadening the potential accessibility of precise and real-time medical assistance.

\end{abstract}

\section{Introduction}
% \textcolor{red}{[CITATIONs]} 
Large language models (LLMs) have demonstrated significant potential in the medical field~\cite{tu2023towards}. 
% \textcolor{red}{[Detail examples]} 
For example, several powerful LLMs have passed medical licensing examinations in various countries, showcasing their capability to solve medical questions on a par with junior doctors~\cite{medical-medpalm}. Consequently, LLMs are being extensively explored in healthcare, ranging from drafting medical reports to assisting clinical decision-making~\cite{thirunavukarasu2023large}.  

% \textcolor{red}{[Challenges in encoding enough clinical knowledge, lack of data from real practice EHR data, online consultation data ]} 

% \textcolor{red}{[Start with impacts and benefits of online medical consultation. Accessible, real-time, low-cost. Then, discuss the application of LLM in this area. The potential of using LLM in this area. ]}

Among all potential areas in medical domain, online medical consultation is probably the most suitable for LLM application. Online medical consultation has the following advantages: 1) it can increase patients’ accessibility to medical care, especially for those in rural areas \cite{llm-lmic-lancet}; 2) it can let patients feel more relaxed than going to hospital offline, which in turn increases the accuracy and completeness of collected main compliant; 3) it can protect the privacy of patients. Due to above advantages, the number of online medical consultations grows at an explosive speed. The outbreak of COVID-19 has further boosted the adoption of online medical consultation. According to a recent statistics~\footnote{https://www.globalmarketestimates.com/market-report/global-online-doctor-consultation-market-2172}, the market of online medical consultation is valued on 3.9 Billion USD in 2020 and is estimated to achieve 16 Billion in 2026.

Specifically, LLMs can improve the efficiency of online medical consultation from two perspectives: 1) for experienced doctors, LLMs can generate the draft response automatically, the doctors only need to modify it rather than start from scratch; 2) for inexperienced doctors, LLMs can reminder the possible examination to take or further inquiries on symptoms, which in turn avoids the misdiagnosis and missed diagnosis. Due to these potentials of LLMs in online medical consultation, growing research efforts have been devoted to this area \cite{llm-consulation-ph-jama, llm-consulation-cardio-jama, llm-chatbot-nejm}. However, there are still two remaining challenges: 1) LLM could produce hallucinations which is not tolerable in medical domain; 2) LLM is now a black box, and the inference procedure is hidden. Therefore, doctors are hard to uncover the chain of thoughts of LLMs. 

To address above two challenges, we introduce the \textbf{Med}ical dialogue with \textbf{K}nowledge enhancement and clinical \textbf{P}athway encoding (\textbf{MedKP}) framework. MedKP consists of two core modules: 1) External Knowledge Enhancement: this module extracts related knowledge from a pre-built medical knowledge graph. The extracted knowledge can help to guide the generation process of LLMs; 2) Internal Clinical Pathway Encoding: this module mines key points from historical conversations and the actions taken by doctors. This mined information ensures the clinical coherence of the entire conversation.

To evaluate whether the proposed MedKP can relieve the hallucination problem, in addition to the common natural language generation metrics like ROUGE \cite{metrics-rouge} and BertScore \cite{metrics-bert}, we also introduce two types of new metrics: 1) entity-based metric which helps to judge whether the key information can accurately be captured; 2) LLM-judge based metrics which evaluate the hallucination of generated responses.
% promising to improve online medical dialogue generation, offering intelligent healthcare in a more accessible manner to serve a wider population.
% However, 
% \textcolor{red}{[Challenges in applying LLM in this area. knowledge gap, ]}
% \textbf{Med}ical dialogue with \textbf{K}nowledge enhancement and clinical \textbf{P}athway encoding (\textbf{MedKP})
Overall, our main contributions are summarized as follows:
\begin{itemize}

\item We propose MedKP which enhances the automatic medical dialogue system with two core modules: External Knowledge Enhancement through a medical knowledge graph, and Internal Clinical Pathway Encoding via medical entities and physician actions.
\item Integrating these enhancements with a generative Large Language Model (LLM) for online medical consultations significantly reduces the hallucinations.
% typically observed in LLMs.
% and substantially enhances the model's interpretability.
\item MedKP outperforms baseline models across two datasets, achieving state-of-the-art results. Comprehensive ablation studies underscore the contribution of individual components to the overall efficacy of our approach.

\end{itemize}

\begin{figure*}
    \centering
        \includegraphics[width=0.95\linewidth]{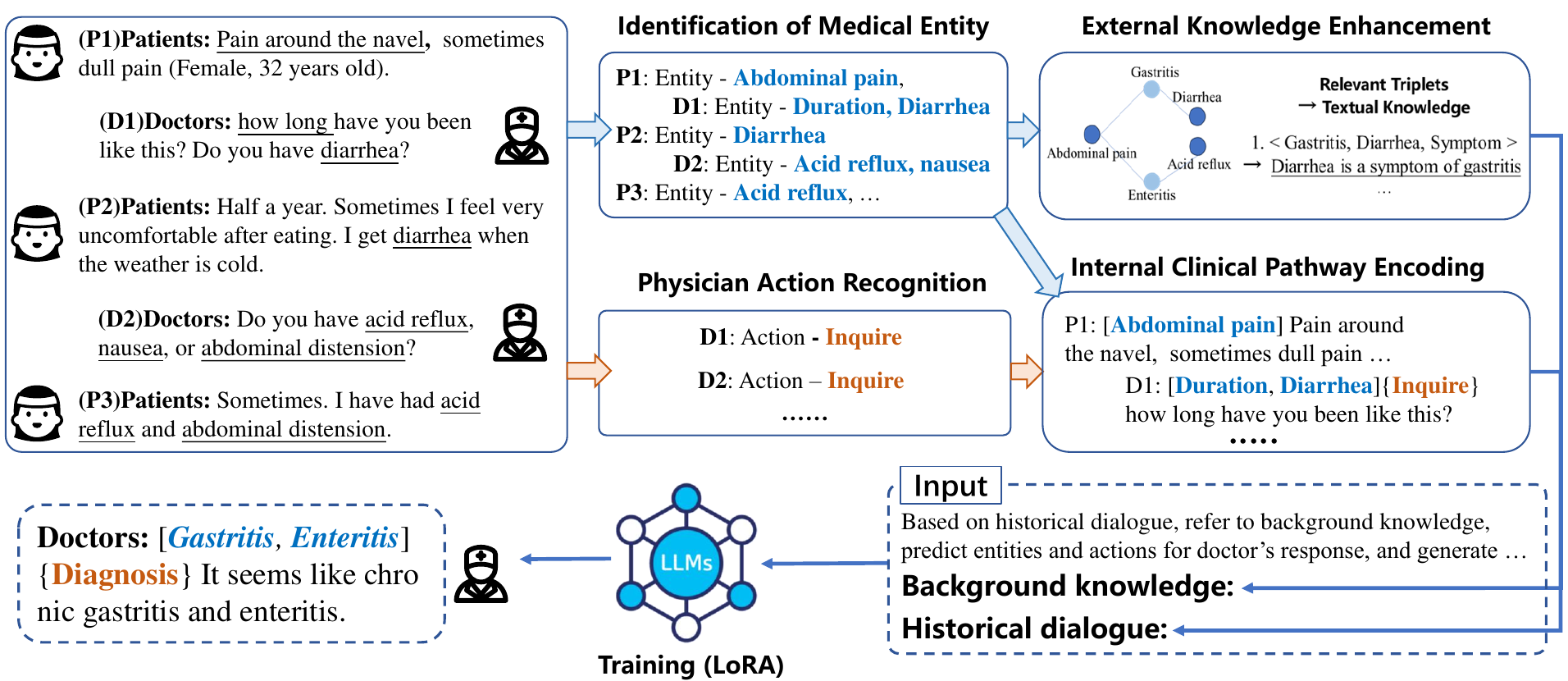}
    \caption{Workflow of medical dialogue with knowledge enhancement and clinical pathway encoding framework.}
    \label{fig:workflow}
\end{figure*}
% \footnote{All prompts and medical dialogue in this study were written in Chinese to be consistent with the actual dataset, we translated them into English in the paper for better readability}

\section{Related Work}
\subsection{Large Language Model in Healthcare}
Owing to extensive pre-training, LLMs encapsulate a broad spectrum of medical knowledge \cite{llm-med-natmed}. 
For general LLMs, GPT-4 \cite{medical-gpt4-2023} surpassed the USMLE passing score by more than 20 points.
For medical-specific LLMs, Med-PaLM \cite{medical-medpalm} and Med-PaLM 2 \cite{medical-medpalm-2}, achieved high scores of 67.6\% and 86.5\% on USMLE respectively, indicating their expert-level proficiency in handling medical questions. 
Additionally, \citet{cnmle-chatgpt-2023} quantified Chinese Medical Licensing Examination by knowledge-enhanced LLMs.

There are emerging studies devoted to applying LLMs in the medical domain. \citet{llm-simplify-chatgpt} and \citet{llm-simplify-gpt4} leverage ChatGPT and GPT-4 to translate radiology reports into plain language. ChatCAD \cite{llm-chatcad} incorporates the LLMs for an interactive computer-aided diagnosis of radiology images.
However, applying LLM in the healthcare domain also raised concerns over the generations of hallucination \cite{medical-hallucination-2023} or biased results \cite{eye-llm-2024}.

\subsection{Medical Dialogue System}
Medical dialogue systems aim to automatically generate responses to patient inquiries, streamlining the delivery of medical services \cite{review-medical-dialogue}. 
For disease diagnosis, \citet{dialogue-diagnosis-muzhi} and \citet{medical-dialogue-meddg} have developed systems for symptom collection and diagnosis using task-oriented dialogues and knowledge graphs, respectively.
For general responses, \citet{medical-dialogue-meddg} and \citet{medical-dialogue-vrbot} focus on entity-driven dialogue generation for more accurate responses. Plugmed \cite{baseline-plugmed} exploits LLMs' in-context learning for generating physician responses. 
While studies have integrated medical entity or knowledge graphs, they often employ additional models for entity prediction or encoding, leading to a lack of interpretability and fragmented processes that may omit crucial details, like symptom states (positive/negative).

\section{Methodology}

\subsection{Problem Formulation}
Each medical dialogue consists of inquiries from the patient and responses from the physician, which we define as $U=\{P, D\}$ to represent a whole medical dialogue. Here, $P=\{p_1, p_2, ..., p_n\}$ denotes the patient's utterances, while $D=\{d_1, d_2, ..., d_{n-1}\}$ represents the physician's utterances. The dialogue between patient and physician alternates in chronological order. An automated medical dialogue system aims to generate a physician's response $d_n$ automatically, based on $U$ the patient-physician dialogue up to the moment $n$ and the current patient inquiry $p_n$, thereby completing the response. 

\subsection{Overall Workflow}
To enhance the reliability of automatically generated responses, we introduce the \textbf{Med}ical dialogue with \textbf{K}nowledge enhancement and clinical \textbf{P}athway encoding (\textbf{MedKP}) framework, the whole workflow of which is illustrated in Figure \ref{fig:workflow}. This framework comprises three main components: 
\begin{itemize}
    \item External Knowledge Enhancement: This module identifies medical entities previously mentioned in historical utterances and retrieves relevant knowledge from the medical knowledge graph. This process enriches the dialogue with reliable medical knowledge.
    \item Internal Clinical Pathway Encoding: It encodes the clinical pathway contained in historical dialogue using medical entities and physician actions. This aids in capturing the medical information conveyed in past conversations and understanding the current state, thereby ensuring a coherent and informed progression of the medical dialogue. 
    \item Response Generation: The relevant medical knowledge and encoded historical utterances are formatted with a specified prompt template to leverage the in-context learning ability of LLMs. We further fine-tuned LLM with the LoRA framework to augment its ability to utilize external medical knowledge and internal clinical pathways. 
    
\end{itemize}
Furthermore, we design a comprehensive automatic evaluation scheme to assess the response quality, including metrics related to medical entities, Natural Language Generation (NLG), and judgment of hallucination based on LLM.

\subsection{External Knowledge Enhancement}
In this section, we explore how to integrate the knowledge graph to mine reliable medical knowledge, serving as the knowledge foundation of medical dialogue generation. 
Initially, we identify medical entities contained in each utterance by patients and physicians, including symptoms, drugs, examinations, and diseases. We aggregate all entities from historical utterances up to the current turns $n$, representing as $E=\{e_1, e_2, ..., e_m\}$. We incorporate a large-scale medical knowledge graph $G=\{K, T\}$, where $K$ signifies all nodes and $T$ represents all triplets in $G$. Each triplet is represented as $t_{ij}=<k_{i}, k_{j}, r_{ij}>$, with $k_i$ as the head node, the $k_j$ as the tail node, and $r_{ij}$ denoting their relationship.

\paragraph{Direct knowledge among mentioned entities}
We first identify the interrelationships existing among the previously referenced medical entities, establishing a direct knowledge foundation. 
For each entity $e_i$ within the entity set $E$, we explore connections with the other entities in $E$, aiming to identify pair $\{e_i, e_j\} \in E$ where a triplet $t_{ij}$ in the knowledge graph denoting the relationship between $e_i$ and $e_j$. This process enables the construction of $T_{direct}$, defined as:
\begin{equation}
T_{direct} = \{t_{ij} \, | \,e_i, e_j \in E \, and \, t_{ij} \in T\}
\end{equation}

\paragraph{Potential knowledge from related entities} 
By integrating the interrelationships among medical entities and the network structure of the knowledge graph, we can also mine medical concepts that are not yet present in historical dialogue but are significantly related, serving as potential knowledge supplements. 
Initially, we retrieve all nodes $K_E$ and edges $R_E$ from G related to the current entity set $E$. Subsequently, we identify nodes not in $E$ but frequently connected to entities within $E$. 
Specifically, for each $k \in K_E$, we calculate the frequency of its relation to entities in $E$, selecting the top-N nodes most related to multiple entities in $E$ as potential co-related nodes $K_{potential}$, formulated as:
\begin{equation}
\small
K_{potential}=top\text{-}N\{k|k \in K_E, max {\sum_{e_i \in E}}freq(k, e_i) \}
\end{equation}
Here, $freq(k, e_i)$ denotes the frequency of the relationship between node $k$ and entity $e_i \in E$.
Further, we extract triplet set $T_E$ that is related to $K_{potential}$, considering them as potential knowledge supplement $T_{potential}$. The process is defined as:
\begin{equation}
\small
T_{potential} = \{t_{jk}| e_k \in K_{potential}, e_j \in E, t_{jk} \in T_E \}
\end{equation}
It systematically identifies and incorporates potentially relevant medical concepts, enriching the context with unexplored but significant knowledge. 

\subsection{Internal Clinical Pathway Encoding}
% Different from conventional question-and-answer tasks, automatic medical dialogue systems represent a dynamic multi-turn dialogue where physicians not only address patients' queries but also guide patients to state their detailed symptoms through inquiry, thereby offering concrete diagnostic and treatment recommendations. 
To accurately represent the dynamic state of medical dialogue, we mine medical entities together with physician actions, thereby encoding the underlying clinical pathways. 
We identify medical entities within each patient and doctor utterance, labeled as $E_{p_n}$ and $E_{d_n}$ respectively. 
Furthermore, we analyze each physician's response $d_n$ to recognize its $s$ actions into $A_{d_n}=\{a_1,a_2,...,a_s\}$, with each $a$ belonging to a predefined set of actions $A$. 
Following previous studies \citet{medical-dialogue-vrbot} and \citet{medical-dialogue-dfmed}, we employ the SOAP note framework \cite{action-soap}, a widely used method of documentation for physicians, to define seven types of physician actions $A$: \textit{Chitchat}, \textit{Inform}, \textit{Inquire}, \textit{Provide Daily Precaution}, \textit{State a Required Medical Test}, \textit{Make a Diagnosis}, and \textit{Prescribe Medications}.

Amidst this encoding scheme for the clinical pathway, we concatenate each utterance with its identified entities and actions. The patient's utterance is encoded as:
\begin{equation}
p'= (E_{p} \parallel p)
\end{equation}
where $E_{p}$ denotes the entities identified in the patient's utterance, and $p$ is the utterance text. Similarly, the physician's utterance is encoded as:
\begin{equation}
d'= (E_{d} \parallel A_{d} \parallel d)
\end{equation}

\subsection{Response Generation}
\subsubsection{Inference}
To generate the response with knowledge enhancement and encoded clinical pathways, we employ a prompt template to format our input for LLM. As depicted in Figure \ref{fig:workflow}, we instruct the LLM with a detailed description following the relevant knowledge and encode historical utterances. 

% To generate the physician's response $d_n$ with a generative LLM, the average negative log-likelihood of the target sequence $d_t=\{w_1, w_2, ..., w_l\}$ is conventionally used as the generation loss:
% \begin{equation}
% \begin{split}
% \mathcal{L}_{g}=-\frac{1}{l}\sum_{l=1}^{L}logP(w_l)
% \end{split}
% \end{equation}

\subsubsection{Training}
Given the encoded historical dialogue $U'=\{P', D'\}$, the model is firstly tasked with generating the entities $e_{d_n}$ to be involved in the response and the action $A_{d_n}$, followed by the actual textual response.
The objective of the generative language model is formalized to maximize the probability of generating the physician's response $d_n$, which can be represented as:
\begin{equation}
\mathop{\arg\max}\limits_{\theta} P(d_n \,| \, U', e_{d_n}, A_{d_n})
\end{equation}
where $\theta$ denotes the current model parameters. With the incorporation of entity and action information, the loss function is enriched to not only account for the accuracy of the generated text but also the relevance and correctness of the entities and actions. Hence, the loss function $\mathcal{L}$ can be formulated as:
\begin{equation}
\begin{split}
\mathcal{L}=&\mathcal{L}_{g}(E_{d_t}, \hat{E_{d_t}}) + \mathcal{L}_{g}(A_{d_t}, \hat{A_{d_t}})
\\
&+\mathcal{L}_{g}(d_t, \hat{d_t})
\end{split}
\end{equation}
\begin{equation}
\begin{split}
\mathcal{L}_{g}(y, \hat{y})=-\frac{1}{L}\sum_{l=1}^{L}y_l \log \hat{y_l}
\end{split}
\end{equation}
where $\mathcal{L}_{g}(E_{d_t}, \hat{E_{d_t}})$ penalizes discrepancies between the predicted and actual entities and $\mathcal{L}_{g}(A_{d_t}, \hat{A_{d_t}})$ assesses the accuracy of the predicted actions against the predefined set. These components of the loss function synergistically guide the response generation, ensuring that the output not only aligns with the factual content but also adheres to the appropriate actions, thereby enhancing the reliability of the generated response.

% \begin{equation}
% \begin{split}
% \mathop{\arg\max}\limits_{\theta} \, P( d_t | E_{d_t}, A_{d_t}, U_{t-1}, p_t ) 
% \end{split}
% \end{equation}

% \begin{table}[!tp]
%     \centering
%     \tiny
%     \caption{Statistics of MedDG and KaMed}
%     \resizebox{\linewidth}{!}{
%     \begin{tabular}{ccccc}
%     \toprule
%        \multirow{2}{*}{\textbf{Dataset}} & \multirow{2}{*}{\textbf{Train/Valid/Test}} & \multirow{2}{*}{\textbf{\begin{tabular}[c]{@{}c@{}}Number of \\ entity type\end{tabular}}} & \multirow{2}{*}{\textbf{\begin{tabular}[c]{@{}c@{}}Average turn of \\ each dialogue\end{tabular}}} & \multirow{2}{*}{\textbf{\begin{tabular}[c]{@{}c@{}}Physician's response with \\ medical entity ($\geq 1$)\end{tabular}}} \\
%         &  &  &  &  \\
%     \midrule
%         MedDG & 14,862/1,999/999 & 160 & 11.62 & 53,984/7,205/3,835 \\
%         KaMed & 29,159/1,532/1,539 & 5,862 & 9.92 & 171,038/9,109/9,159 \\
%     \bottomrule
%     \end{tabular}
%     }
%     \label{table:dataset}
% \end{table}
% \subsection{Dataset}

To expedite the training of LLMs, we employ the LoRA framework \cite{lora-2021} to implement parameter-efficient fine-tuning. 
By freezing the parameters of the base LLM and integrating additional LoRA layers specifically for training, we can effectively tailor the model.
This strategy enables efficient adaptation of the LLM to incorporate external knowledge and clinical pathways, providing more reliable responses.

\subsection{Evaluation}
Previous studies \cite{survey-hallucination-2023, metrics-qa-2021} have observed that conventional natural language generation (NLG) metrics face challenges in efficiently measuring the quality of open-domain text generation tasks. 
Especially in the medical domain, relying solely on character overlap without considering the actual semantics and the information conveyed fails to accurately and objectively evaluate the quality of text generation \cite{metrics-medical-report}.
Therefore, we employ comprehensive metrics to assess the quality of generated responses and whether our method alleviates the hallucinations. 

\begin{figure}[!tp]
    \centering
        \includegraphics[width=7.8cm]{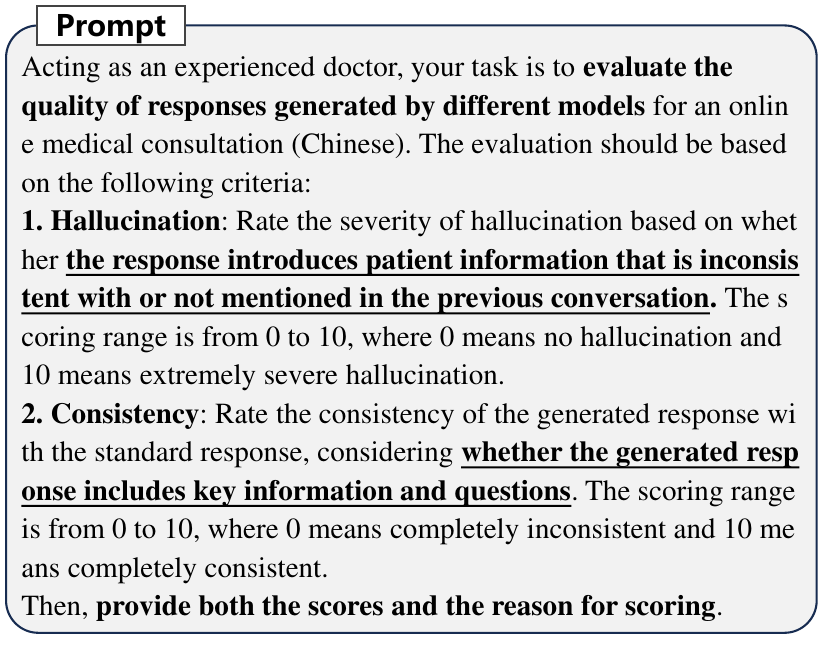}
    \caption{Prompt for LLM judge.}
    \label{fig:prompt of llm-judge}
\end{figure} 

\paragraph{NLG metrics} We adopt the ROUGE \cite{metrics-rouge} and BLEU \cite{metrics-bleu} to evaluate the quality of generated responses at the character level.
% , which are commonly applied in the field of medical text generation. 
Specifically, we utilize BLEU-1/2/3/4 and ROUGE-1/2/L to measure in different n-grams.
% The former reflects the recall of n-grams from the ground truth in the generated text, while the latter reflects the precision of n-grams in the generated text that appears in the ground truth. 

\paragraph{Text similarity} We use the BertScore \cite{metrics-bert} to measure the overall similarity between the generated text and the target text, primarily leveraging BERT to compute the semantic distance between texts.

\paragraph{Medical Entity} To better evaluate the accuracy of generated responses in medical contexts, we assess their performance at the entity level. Specifically, we calculate recall, precision, and F1-score for entities that should be mentioned in the responses. Moreover, while previous studies often adopted only micro-metrics following \citet{medical-dialogue-meddg}, this approach may overlook the accuracy of sentences that are shorter or contain fewer entities. Consequently, we calculate both macro- and micro-metrics to provide a comprehensive assessment.

\paragraph{LLM judge} The previous metrics only evaluate the differences between generated responses and corresponding ground truth, neglecting the contextual background of historical dialogue. 
To address this, we have leveraged the in-context learning capabilities of advanced models to construct a judge based on LLMs \cite{metric-llm-judge}. This judge primarily focuses on consistency with previous context (to mitigate hallucination) and consistency with subsequent responses (to ensure consistency). Specifically, as illustrated in Figure \ref{fig:prompt of llm-judge}, we have designed a template for GPT-4 to act as an experienced doctor evaluating the quality of generated responses. 
The evaluation is based on the following criteria, with scoring and reasoning provided to enhance interpretability:

\textbf{Hallucination} (0-10): Measures whether the response introduces information that conflicts with or is not mentioned in the preceding text. A lower score indicates fewer hallucinations.

\textbf{Consistency} (0-10): Assess whether the response aligns with subsequent physician responses, including key information and questions. A higher score indicates stronger consistency.

\begin{table*}[!tp]
    \centering
    \scriptsize
    \caption{Performance evaluation of medical entities and LLM judge on MedDG. Rec/Pre/F1 stand for Recall/Precision/F1-score, which together evaluate the accuracy of predicting medical entities, H represents Hallucination, assessing the generation of non-factual information, and C stands for consistency, evaluating the logical coherence between ground-truth and the generated one.}
    % \resizebox{\linewidth}{!}{
    \begin{tabular}{clcccccccc}
    \toprule
    \multicolumn{2}{c}{\multirow{2}{*}{\textbf{Method}}} & \multicolumn{3}{c}{\textbf{Medical Entity-macro}} & \multicolumn{3}{c}{\textbf{Medical Entity-micro}} & \multicolumn{2}{c}{\textbf{LLM-Judge}} \\ \cline{3-10} 
    \multicolumn{2}{c}{} & \textbf{Rec} & \textbf{Pre} & \textbf{F1} & \textbf{Rec} & \textbf{Pre} & \textbf{F1} & \textbf{H} & \textbf{C} \\
    % \cline(3-10)
    \midrule
    \multirow{5}{*}{\textbf{DL-based}} & Seq2Seq & 13.49 & 15.78 & 13.98 & 10.42 & 27.18 & 15.07 & 1.80 & 3.76 \\
     & Seq2Seq-Entity & 19.42 & 21.86 & 19.27 & 15.91 & 35.79 & 22.03 & 1.13 & 4.16 \\
     & HRED & 12.87 & 15.18 & 13.29 & 10.03 & 25.55 & 14.40 & 1.71 & 3.52 \\
     & HRED-Entity & 19.01 & 20.96 & 18.63 & 15.41 & 33.02 & 21.01 & 1.64 & 3.86 \\
     & VRBOT & 11.90 & 14.99 & 12.56 & 9.49 & 29.31 & 14.34 & 1.53 & 3.47 \\
     \midrule
    \multirow{5}{*}{\textbf{PLM-based}} & GPT-2 & 17.13 & 19.62 & 17.27 & 14.34 & 29.19 & 19.23 & 1.05 & 4.30 \\
     & GPT-Entity & 20.06 & 22.71 & 19.96 & 16.99 & 32.12 & 22.22 & \textbf{1.03} & 4.67 \\
     & BART & 17.53 & 20.58 & 17.89 & 14.28 & 30.83 & 19.52 & 1.05 & 4.54 \\
     & BART-Entity & 20.76 & 22.43 & 19.92 & 16.98 & 35.56 & 22.98 & 1.11 & 4.54 \\
     & DFMed & 27.98 & 26.14 & 24.76 & 24.13 & 32.45 & 27.68 & 1.06 & 5.39 \\
     \midrule
    \multirow{2}{*}{\textbf{LLM-based}} & Direct Inference & 13.65 & 13.71 & 12.41 & 12.62 & 17.00 & 14.49 & 2.60 & 3.49 \\
    % & PlugMed & - & - & - & - & - & - & - & - \\
     & MedKP & \textbf{32.38} & \textbf{35.11} & \textbf{31.41} & \textbf{28.12} & \textbf{29.62} & \textbf{28.85} & \textbf{1.03} & \textbf{6.10} \\
     \bottomrule
    \end{tabular}
    \label{table:performance of ent in meddg}
    % }
\end{table*}

\section{Experiments}
We conduct experiments on two public datasets of medical dialogue for our evaluation:
% , collected from real-world medical consultation websites after anonymization, thereby mirroring clinically authentic medical scenarios closely:
(1) \textbf{MedDG} dataset \cite{medical-dialogue-meddg} is sourced from \textit{Doctor Chunyu}\footnote{https://www.chunyuyisheng.com/}. It comprises 17,684 medical dialogues, primarily focusing on 12 gastrointestinal diseases. This entity-centric dataset is systematically annotated by physicians and defines 160 normalized medical entities across five types: disease, symptom, medicine, examination, and attribute. Adhering to the official dataset partition, we divide the dataset into 14,862/1,999/999 dialogues for the training, validation, and test sets, respectively. 
(2) \textbf{KaMed} dataset \cite{medical-dialogue-vrbot} is also derived from \textit{Doctor Chunyu}, caters to diverse clinical scenarios with its inclusion of over 100 hospital departments (No overlap with MedDG). It is larger than previous datasets in scale and also features more rounds of conversation, making it more challenging.
Additionally, the original sessions of KaMed contain multi-modal information such as images and voice recordings, which have been replaced by the meaningless template "The image/voice is not available for privacy concern", leading to incomplete information. Following the filtering rules established by \citep{medical-dialogue-dfmed}, we have filtered the raw dataset to curated 29,159/1,532/1,539 for training/validation/testing.

\begin{table*}[!tp]
    \centering
    \scriptsize
    \caption{Performance evaluation of NLG metrics and text similarity on MedDG}
    % \small
    % \resizebox{\linewidth}{!}{
    \begin{tabular}{clcccccccc}
    \toprule
    \multicolumn{2}{c}{\multirow{2}{*}{\textbf{Method}}} & \multicolumn{3}{c}{\textbf{NLG-ROUGE}} & \multicolumn{4}{c}{\textbf{NLG-BLEU}} & \textbf{Text Similarity} \\ \cline{3-10} 
    \multicolumn{2}{c}{} & \textbf{R-1} & \textbf{R-2} & \textbf{R-L} & \textbf{B-1} & \textbf{B-2} & \textbf{B-3} & \textbf{B-4} & \textbf{BertScore} \\
    \midrule
    \multirow{5}{*}{\textbf{DL-based}} & Seq2Seq & 21.90 & 9.13 & 20.95 & 21.86 & 17.33 & 14.55 & 11.84 & 63.35 \\
     & Seq2Seq-Entity & 22.74 & 9.60 & 21.49 & 23.49 & 18.49 & 15.51 & 12.60 & 64.02 \\
     & HRED & 21.72 & 8.88 & 20.48 & 25.56 & 20.42 & 17.34 & 14.22 & 63.47 \\
     & HRED-Entity & 22.55 & 9.03 & 20.99 & 27.63 & 21.78 & 18.36 & 14.86 & 63.87 \\
     & VRBOT & 20.41 & 8.59 & 19.42 & 22.89 & 18.24 & 15.52 & 12.71 & 62.30 \\
     \midrule
    \multirow{5}{*}{\textbf{PLM-based}} & GPT-2 & 25.05 & 11.15 & 23.50 & 28.27 & 22.28 & 18.75 & 15.28 & 65.00 \\
     & GPT-Entity & 25.51 & 11.30 & 23.79 & 28.31 & 22.21 & 18.62 & 15.13 & 65.20 \\
     & BART & 25.37 & 11.50 & 23.85 & 27.32 & 21.49 & 17.97 & 14.53 & 65.20 \\
     & BART-Entity & 24.99 & 11.19 & 23.37 & 27.22 & 21.33 & 17.88 & 14.53 & 65.07 \\
     & DFMed & 28.22 & 12.81 & 25.07 & \textbf{38.93} & 29.81 & 24.82 & 20.00 & 66.58 \\
     \midrule
    \multirow{3}{*}{\textbf{LLM-based}} & PlugMed & - & - & 21.10 & - & - & - & - & 64.10 \\
     & Direct Inference & 18.51 & 5.09 & 15.55 & 37.83 & \textbf{30.09} & \textbf{25.70} & \textbf{20.59} & 60.70 \\
     & MedKP & \textbf{29.50} & \textbf{14.25} & \textbf{26.86} & 37.41 & 29.08 & 24.24 & 19.64 & \textbf{67.35} \\
     \bottomrule
    \end{tabular}
    \label{table:performance of nlg in meddg}
    % }
\end{table*}

% Table \ref{table:dataset} presents the statistics of these datasets. 

Notably, real medical consultations often contain many simple sentences, such as chitchat or greetings, with few words and lacking medical information. To efficiently develop a robust automatic medical dialogue generation system, our evaluation primarily focuses on the challenging responses that contain at least one medical entity.

\begin{table*}[htbp]
    \centering
    \scriptsize
    \caption{Performance evaluation of medical entities and LLM judge on KaMed.}
    % \small
    % \resizebox{\linewidth}{!}{
    \begin{tabular}{clcccccccc}
    \toprule
        \multicolumn{2}{c}{\multirow{2}{*}{\textbf{Model}}} & \multicolumn{3}{c}{\textbf{Medical Entity-macro}} & \multicolumn{3}{c}{\textbf{Medical Entity-micro}} & \multicolumn{2}{c}{\textbf{LLM-Judge}} \\ 
        \cline{3-10} 
        \multicolumn{2}{c}{} & \multicolumn{1}{c}{\textbf{Rec}} & \textbf{Pre} & \textbf{F1} & \textbf{Rec} & \textbf{Pre} & \textbf{F1} & \textbf{H} & \textbf{C} \\
    \midrule
        \multirow{3}{*}{\textbf{DL-based}} & Seq2Seq & 8.70 & 10.13 & 8.91 & 8.48 & 18.93 & 11.72 & 2.32 & 1.72 \\
         & HRED & 8.94 & 10.26 & 9.11 & 8.40 & 15.62 & 10.93 & 1.81 & 1.44 \\
         & VRBOT & 6.18 & 7.64 & 6.56 & 5.87 & 16.17 & 8.61 & 2.28 & 1.66 \\
     \midrule
        \multirow{3}{*}{\textbf{PLM-based}} & GPT-2 & 14.98 & 16.03 & 14.67 & 13.70 & 21.13 & 16.62 & 0.80 & 2.98 \\
         & BART & 16.50 & 17.56 & 16.20 & 15.40 & 23.69 & 18.66 & 0.71 & 3.23 \\
         & DFMed & 27.84 & 26.62 & 25.75 & 25.45 & 23.37 & 24.37 & 0.69 & 4.04 \\
     \midrule
        \multirow{2}{*}{\textbf{LLM-based}} & Direct Inference & \multicolumn{1}{c}{20.61} & 21.01 & 19.74 & 19.11 & 23.57 & 21.11 & 0.96 & 4.10 \\
        % & PlugMed & \multicolumn{1}{c}{-} & - & - & - & - & - & - & - \\
         & MedKP & \textbf{35.25} & \textbf{32.49} & \textbf{31.71} & \textbf{33.09} & \textbf{24.12} & \textbf{27.90} & \textbf{0.20} & \textbf{5.38} \\
    \bottomrule
    \end{tabular}   
    \label{table:performance of ent in kamed}
    % }
\end{table*}

\subsection{Baseline models}
To fully evaluate the performance of different methods in medical dialogue generation, we constructed several baselines that cover deep learning(DL)-based methods, pre-trained language model(PLM)-based methods, and LLM-based methods.
% \begin{itemize}
    % \item 
    
    \textbf{DL-based method}: 
    (1) Seq2Seq \cite{baseline-seq2seq} is a classical sequence to sequence model, employing an attention mechanism coupled with RNN-based architectures for both the encoder and decoder components. 
    (2) HRED \cite{baseline-hred} advanced the conventional Seq2Seq encoder by employing a hierarchical structure that models a dialogue as a token sequence and an utterance sequence. 
    (3) VRBot \cite{medical-dialogue-vrbot} is an end-to-end variational reasoning model for medical dialogue generation that tracks patient state and physician action. 
    % \item 

    \textbf{PLM-based method}: 
    (1) GPT-2 \cite{baseline-gpt2} is a classical transformer-decoder-based language model.
    (2) BART \cite{baseline-bart} is a transformer-based encoder-decoder model. 
    (3) DFMed \cite{medical-dialogue-dfmed} employs two sequential models to predict medical entities and physician actions, respectively. It introduces an interweaving component designed to integrate those predicted states to generate a physician's response.
    % \item 
    
    \textbf{LLM-based method}: (1) PlugMed \cite{baseline-plugmed} retrieves similar dialogues to guide LLMs to generate responses and fine-tune a small model to discern the best responses. They employ BLOOM as the foundation model. Due to its code and dataset is not available, we only record and compare the reported metrics.
    (2) Direct Inference represents the direct application of ChatGLM3-6B\footnote{https://github.com/THUDM/ChatGLM3} for response generation without enhancement or fine-tuning
% \end{itemize}
Furthermore, leveraging the high-quality entities defined by physicians within the MedDG dataset, we have also augmented various baselines with entity enhancement. Following \citep{medical-dialogue-meddg}, we entail appending entities predicted by an auxiliary model directly to the dialogue history. Such augmentation serves as a hint for the generative language model.

\subsection{Implementation Details}

% Notably, it achieved the best result for various Chinese tasks among LLMs with fewer than 10B parameters. 

% \footnote{https://github.com/huggingface/peft}
% \footnote{https://github.com/microsoft/DeepSpeed} 
% For fine-tuning the LLM, we employed LoRA with the PEFT library(version 0.6.0). 
% All experiments are conducted on 8*Nvidia Tesla V100 GPUs, utilizing the DeepSpeed framework for parallel training and acceleration. 

% The configurations for training are summarized as follows:
% % \begin{itemize}
% %     \item 
    
%     \textbf{LoRA parameters}:
   We select ChatGLM3-6B as our base LLM.
   % , an open-source model comprising 6B parameters, that can be conveniently deployed on common PCs.  
    The adaptation utilized a LoRA rank $r$ of 8, scaling factor $\alpha$ of 32, and dropout rate of 0.1. The layers designated for training within the architecture of ChatGLM3 include the self-attention components and linear layers, specifically: "query\_key\_value", "dense", "dense\_h\_to\_4h", "dense\_4h\_to\_h"
    % \item 
    % \textbf{Training setting}: 
    We set the batch size to 64, conducting the training over 20 epochs for each dataset. The AdamW optimizer starts with a learning rate of 5e-4 and decreases to 5e-5. To satisfy $99\%$ of data, the maximum input length is 1,536 tokens and the maximum output length is 256 tokens.
% \end{itemize}

% For knowledge enhancement, 
We select CMEKG\footnote{http://cmekg.pcl.ac.cn/} as the knowledge graph and select top 5 commonly related entities for mining the potential knowledge.
For all baselines, we implement the open-source code and follow their settings provided by \citet{medical-dialogue-meddg}, \citet{medical-dialogue-vrbot}, and \citet{medical-dialogue-dfmed}. The MedBERT\footnote{https://github.com/trueto/medbert} pretrained in the medical domain is selected as the backbone of PLM-based methods.
For LLM Judge, we conduct tests by calling OpenAI's official API with the model version `GPT4-0125-preview'. Due to the API access rate limitations, we randomly selected 500 samples in each dataset for testing.

\begin{table*}[htbp]
    \centering
    \scriptsize
    \caption{Performance evaluation of NLG metrics and text similarity on KaMed.}
    % \resizebox{\linewidth}{!}{
    \begin{tabular}{clcccccccc}
    \toprule
        \multicolumn{2}{c}{\multirow{2}{*}{\textbf{Model}}} & \multicolumn{3}{c}{\textbf{NLG-ROUGE}} & \multicolumn{4}{c}{\textbf{NLG-BLEU}} & \textbf{Text Similarity} \\ 
    \cline{3-10} 
        \multicolumn{2}{c}{} & \textbf{R-1} & \textbf{R-2} & \textbf{R-L} & \textbf{B-1} & \textbf{B-2} & \textbf{B-3} & \textbf{B-4} & \textbf{BertScore} \\
    \midrule
        \multirow{3}{*}{\textbf{DL-based}} & Seq2Seq & 17.85 & 4.71 & 16.96 & 14.16 & 10.99 & 9.14 & 7.23 & 59.86 \\
         & HRED & 18.35 & 4.81 & 17.02 & 18.59 & 14.53 & 12.17 & 9.66 & 60.53 \\
         & VRBOT & 16.98 & 5.60 & 14.85 & 27.18 & 21.38 & 18.38 & 14.85 & 57.43 \\
    \midrule
        \multirow{3}{*}{\textbf{PLM-based}} & GPT-2 & 22.79 & 7.49 & 20.53 & 26.98 & 20.81 & 17.43 & 13.92 & 62.39 \\
         & BART & 23.59 & 8.09 & 21.29 & 26.27 & 20.19 & 16.82 & 13.41 & 62.95 \\
         & DFMed & 26.36 & 9.82 & 22.54 & 36.35 & 27.37 & 22.64 & 17.99 & 64.42 \\
    \midrule
        \multirow{3}{*}{\textbf{LLM-based}} & PlugMed & - & - & 14.10 & - & - & - & - & 60.10 \\
        & Direct Inference & 23.25 & 7.49 & 19.60 & 37.94 & \textbf{29.21} & \textbf{24.62} & \textbf{19.67} & 62.47 \\
        & MedKP & \textbf{27.53} & \textbf{10.93} & \textbf{23.75} & \textbf{38.01} & 28.56 & 23.67 & 18.88 & \textbf{64.90} \\
    \bottomrule
    \end{tabular}   
    \label{table:performance of nlg in kamed}
    % }
\end{table*}

\begin{table*}[htbp]
    \centering
    \caption{Experimental results of ablation study on MedDG. The +KG indicates the integration of knowledge graph-based enhancement; the +CP refers to the incorporation of clinical pathways encoding via both medical entities and physician actions, while +entity signifies encoding solely depends on medical entity; MedKP denotes the cooperation of KG and CP.}
    % \small
    \resizebox{\linewidth}{!}{
    \begin{tabular}{lcccccccccccccccc}
    \toprule
    \multicolumn{1}{c}{\multirow{2}{*}{\textbf{Method}}} & \multicolumn{3}{c}{\textbf{Medical Entity-macro}} & \multicolumn{3}{c}{\textbf{Medical Entity-micro}} & \multicolumn{2}{c}{\textbf{LLM-Judge}} & \multicolumn{3}{c}{\textbf{NLG-ROUGE}} & \multicolumn{4}{c}{\textbf{NLG-BLEU}} & \textbf{Similarity} \\
    \cline{2-17} 
    \multicolumn{1}{c}{} & \textbf{Rec} & \textbf{Pre} & \textbf{F1} & \textbf{Rec} & \textbf{Pre} & \textbf{F1} & \textbf{H} & \textbf{C} & \textbf{R-1} & \textbf{R-2} & \textbf{R-L} & \textbf{B-1} & \textbf{B-2} & \textbf{B-3} & \textbf{B-4} & \textbf{BertScore} \\
    \midrule
    \multicolumn{1}{l}{Direct Inference} & 13.65 & 13.71 & 12.41 & 12.62 & 17.00 & 14.49 & 2.60 & 3.49 & 18.51 & 5.09 & 15.55 & \textbf{37.83} & \textbf{30.09} & \textbf{25.70} & \textbf{20.59} & 60.70 \\
    + KG & 28.27 & 31.39 & 27.73 & 24.17 & 28.91 & 26.33 & \textbf{0.97} & 5.52 & 28.89 & 14.30 & 26.83 & 33.15 & 25.99 & 21.77 & 17.69 & \textbf{67.52} \\
    + Entity & 32.20 & 34.90 & 31.13 & 27.81 & 28.44 & 28.13 & 1.01 & 5.51 & 28.86 & 13.87 & 26.61 & 33.67 & 26.26 & 21.90 & 17.70 & 67.51 \\
    + CP & 31.41 & 34.82 & 30.87 & 26.90 & 30.23 & 28.47 & 1.03 & 5.62 & 28.94 & 14.17 & 26.70 & 34.85 & 27.25 & 22.83 & 18.61 & 67.35 \\
    % + KG and Entity encoding & 32.15 & \textbf{35.62} & \textbf{31.42} & 27.61 & 30.80 & \textbf{29.12} & 1.03 & 5.69 & 28.77 & 14.07 & 26.59 & 34.27 & 26.85 & 22.48 & 18.33 & 67.26 \\
    + MedKP & \textbf{32.38} & \textbf{35.11} & \textbf{31.41} & \textbf{28.12} & \textbf{29.62} & \textbf{28.85} & 1.03 & \textbf{6.10} & \textbf{29.50} & \textbf{14.25} & \textbf{26.86} & 37.41 & 29.08 & 24.24 & 19.64 & 67.35 \\
    \bottomrule
    \end{tabular}
    \label{table:ablation in meddg}
    }
\end{table*}

\section{Results and Analysis}
\label{sec:result and analysis}
\subsection{Main result}
Table \ref{table:performance of ent in meddg} and Table \ref{table:performance of nlg in meddg} present a detailed performance evaluation of different methods applied to the MedDG dataset, while Table \ref{table:performance of ent in kamed} and Table \ref{table:performance of nlg in kamed} extend the evaluation to the KaMed dataset. 
Overall, the proposed MedKP framework exhibits a remarkable superiority over competing baselines, yielding new SOTA results across multiple metrics.

% The detailed observations are summarized by the different metrics as follows:
% \paragraph{Medical Entity} 
\textbf{Medical Entity} 
The PLM-based methods demonstrate superior efficacy over DL-based approaches, with the integration of entity hints also augmenting performance. 
Notably, MedKP significantly outperforms all other baselines.
On the MedDG dataset, the LLM equipped with MedKP achieves a substantial increase in performance, with macro-F1 and micro-F1 scores improving dramatically from 12.41 to 31.41 and 14.49 to 28.85, respectively.
Compared to the previous best-performing baseline, MedKP also yielded considerable gains of 6.65 in macro-F1 and 1.17 in micro-F1. 
The pronounced enhancement in macro-metrics underscores MedKP's proficiency in precisely delivering pertinent medical information, highlighting its effectiveness even in concise responses.
These enhancements suggest that the responses generated by MedKP are not only more informative but also closely mirror the physicians' responses, thereby ensuring the effectiveness of medical consultations. 

\textbf{LLM Judge} The integration of medical entities effectively facilitates the understanding of the dialogue state, thereby reducing hallucinations (e.g., Seq2Seq and Seq2Seq-entity). 
Compared to DL-based models, PLM-based methods showed improved performance, which may be attributed to the proficiency of medical LLM in understanding medical text. 
In contrast, LLM applied directly often introduces irrelevant or conflicting patient information and then suffers from severe hallucinations.
Benefiting from reliable medical knowledge and the precise understanding of both historical and current states afforded by pathway encoding, MedKP significantly reduces hallucinations and surpasses other methods by a notable margin, achieving a 0.71 improvement in consistency on MedDG. 

\textbf{NLG Metrics and Text Similarity} 
On the KaMed and MedDG datasets, MedKP achieved the highest ROUGE scores, outperforming the best baseline by 7.14\% on MedDG and 5.36\% on KaMed. 
Similarly, in terms of sentence-level similarity, as measured by BERTScore, MedKP yielded the best result. 
However, regarding BLEU scores, while MedKP's performance was notably high, the highest scores were obtained by directly applying LLM. 
This discrepancy across metrics may be attributed to the LLM's tendency to generate long and general suggestions, leading to high overlaps with standard responses.
However such unfocused responses are meaningless for subsequent consultation. Consequently, metrics like ROUGE that calculate recall of standard responses and medical entity that reflect key information tend to be lower. 
This highlights the potential risks of using traditional NLG metrics for evaluating rigorous medical text generation.

\subsection{Ablation study}
Table \ref{table:ablation in meddg} demonstrates that each component of MedKP significantly enhances its performance on the MedDG dataset. The integration of external medical KG notably increases the reliability of responses, achieving the lowest rates of hallucination and the highest BERTScore. 
Pathway encoding facilitates an understanding of the current state, favoring the prediction of medical entities that should be discussed in subsequent responses. Moreover, the predicted entities and actions guide the LLM towards generating content that is more focused and aligned with physician responses, as demonstrated by improvements in medical entity-metrics and NLG metrics.
Cooperation with all components, MedKP manifests advantages in several metrics, from entity-related to hallucination, underscoring the effectiveness of our framework. 

% \subsection{Case study}

\section{Conclusion}
\label{sec:conclusion}
In this paper, we present MedKP which generates the response from doctors in online medical consultations with LLMs. To alleviate the hallucination problem, on one hand, MedKP introduces an external medical knowledge graph to guide the generation of LLMs; On the other hand, MedKP identifies the key point and physician actions within a conversation which ensures clinical coherence. To evaluate the hallucination problem, we also introduce entity-based and LLM-judge metrics in addition to the common NLG metrics. Experiments on two public benchmarks that demonstrated the effectiveness of MedKP.

\section*{Limitations}
Several potential limitations should be considered for this study. Firstly, the parallel tests were not conducted on more LLMs. This stems from the fact that some advanced LLMs, such as Med-PaLM, have not yet been made available. It is also due to the high computational resources required for fine-tuning LLMs. 
In addition, the detailed responses were not further examined from a professional perspective, which could better evaluate the quality of the generated response. We are currently collaborating with clinical physicians to further this work, and hope to continue refining it in subsequent studies.

\section*{Ethics Statement}
While the medical dialogue involves patient information, all cases have been anonymized, ensuring that no personal information is disclosed. 
Moreover, the primary objective of this study is to investigate the effectiveness of LLM in medical response generation. The results and conclusions will not serve as medical suggestions. Consequently, they do not have any adverse effect on human healthcare. 

% \section*{Acknowledgements}
% None

% \subsection{Appendices}

% Use \verb|\appendix| before any appendix section to switch the section numbering over to letters. See Appendix~\ref{sec:appendix} for an example.

\bibliography{custom}

\begin{thebibliography}{35}
\expandafter\ifx\csname natexlab\endcsname\relax\def\natexlab#1{#1}\fi

\bibitem[{Ayers et~al.(2023)Ayers, Poliak, Dredze, Leas, Zhu, Kelley, Faix, Goodman, Longhurst, Hogarth et~al.}]{llm-consulation-ph-jama}
John~W Ayers, Adam Poliak, Mark Dredze, Eric~C Leas, Zechariah Zhu, Jessica~B Kelley, Dennis~J Faix, Aaron~M Goodman, Christopher~A Longhurst, Michael Hogarth, et~al. 2023.
\newblock Comparing physician and artificial intelligence chatbot responses to patient questions posted to a public social media forum.
\newblock \emph{JAMA internal medicine}.

\bibitem[{Cameron and Turtle-Song(2002)}]{action-soap}
Susan Cameron and Imani Turtle-Song. 2002.
\newblock Learning to write case notes using the soap format.
\newblock \emph{Journal of Counseling \& Development}, 80(3):286--292.

\bibitem[{Chi et~al.(2019)Chi, Choi, Lin, Thompson, Demiris, Laranjo, Dunn, Tong, Jiang, Provoost et~al.}]{review-medical-dialogue}
Nai-Ching Chi, Yong Choi, Shih-Yin Lin, Hilaire Thompson, George Demiris, L~Laranjo, A~Dunn, H~Tong, R~Jiang, S~Provoost, et~al. 2019.
\newblock A systematic review of health dialog systems.
\newblock \emph{Methods of information in medicine}, 58(06):179--193.

\bibitem[{Dou et~al.(2023)Dou, Jin, Jiao, Zhao, Zhao, and Tao}]{baseline-plugmed}
Chengfeng Dou, Zhi Jin, Wenpin Jiao, Haiyan Zhao, Yongqiang Zhao, and Zhengwei Tao. 2023.
\newblock Plugmed: Improving specificity in patient-centered medical dialogue generation using in-context learning.
\newblock In \emph{Findings of the Association for Computational Linguistics: EMNLP 2023}, pages 5050--5066.

\bibitem[{Harrer(2023)}]{medical-hallucination-2023}
Stefan Harrer. 2023.
\newblock Attention is not all you need: the complicated case of ethically using large language models in healthcare and medicine.
\newblock \emph{EBioMedicine}, 90.

\bibitem[{Hu et~al.(2021)Hu, Shen, Wallis, Allen-Zhu, Li, Wang, Wang, and Chen}]{lora-2021}
Edward~J Hu, Yelong Shen, Phillip Wallis, Zeyuan Allen-Zhu, Yuanzhi Li, Shean Wang, Lu~Wang, and Weizhu Chen. 2021.
\newblock Lora: Low-rank adaptation of large language models.
\newblock \emph{arXiv preprint arXiv:2106.09685}.

\bibitem[{Jeblick et~al.(2023)Jeblick, Schachtner, Dexl, Mittermeier, St{\"u}ber, Topalis, Weber, Wesp, Sabel, Ricke et~al.}]{llm-simplify-chatgpt}
Katharina Jeblick, Balthasar Schachtner, Jakob Dexl, Andreas Mittermeier, Anna~Theresa St{\"u}ber, Johanna Topalis, Tobias Weber, Philipp Wesp, Bastian~Oliver Sabel, Jens Ricke, et~al. 2023.
\newblock Chatgpt makes medicine easy to swallow: an exploratory case study on simplified radiology reports.
\newblock \emph{European radiology}, pages 1--9.

\bibitem[{Ji et~al.(2023)Ji, Lee, Frieske, Yu, Su, Xu, Ishii, Bang, Madotto, and Fung}]{survey-hallucination-2023}
Ziwei Ji, Nayeon Lee, Rita Frieske, Tiezheng Yu, Dan Su, Yan Xu, Etsuko Ishii, Ye~Jin Bang, Andrea Madotto, and Pascale Fung. 2023.
\newblock Survey of hallucination in natural language generation.
\newblock \emph{ACM Computing Surveys}, 55(12):1--38.

\bibitem[{Lee et~al.(2023)Lee, Bubeck, and Petro}]{llm-chatbot-nejm}
Peter Lee, Sebastien Bubeck, and Joseph Petro. 2023.
\newblock Benefits, limits, and risks of gpt-4 as an ai chatbot for medicine.
\newblock \emph{New England Journal of Medicine}, 388(13):1233--1239.

\bibitem[{Lewis et~al.(2019)Lewis, Liu, Goyal, Ghazvininejad, Mohamed, Levy, Stoyanov, and Zettlemoyer}]{baseline-bart}
Mike Lewis, Yinhan Liu, Naman Goyal, Marjan Ghazvininejad, Abdelrahman Mohamed, Omer Levy, Ves Stoyanov, and Luke Zettlemoyer. 2019.
\newblock Bart: Denoising sequence-to-sequence pre-training for natural language generation, translation, and comprehension.
\newblock \emph{arXiv preprint arXiv:1910.13461}.

\bibitem[{Li et~al.(2021)Li, Ren, Ren, Chen, Fan, Ma, and de~Rijke}]{medical-dialogue-vrbot}
Dongdong Li, Zhaochun Ren, Pengjie Ren, Zhumin Chen, Miao Fan, Jun Ma, and Maarten de~Rijke. 2021.
\newblock Semi-supervised variational reasoning for medical dialogue generation.
\newblock In \emph{Proceedings of the 44th International ACM SIGIR Conference on Research and Development in Information Retrieval}, pages 544--554.

\bibitem[{Lin(2004)}]{metrics-rouge}
Chin-Yew Lin. 2004.
\newblock Rouge: A package for automatic evaluation of summaries.
\newblock In \emph{Text summarization branches out}, pages 74--81.

\bibitem[{Liu et~al.(2022)Liu, Tang, Cheng, Li, Zheng, and Liang}]{medical-dialogue-meddg}
Wenge Liu, Jianheng Tang, Yi~Cheng, Wenjie Li, Yefeng Zheng, and Xiaodan Liang. 2022.
\newblock Meddg: an entity-centric medical consultation dataset for entity-aware medical dialogue generation.
\newblock In \emph{CCF International Conference on Natural Language Processing and Chinese Computing}, pages 447--459. Springer.

\bibitem[{Liu et~al.(2024)Liu, Wu, Shao, Shen, Ye, Wang, Ye, Jin, Yang et~al.}]{eye-llm-2024}
Xiaocong Liu, Jiageng Wu, An~Shao, Wenyue Shen, Panpan Ye, Yao Wang, Juan Ye, Kai Jin, Jie Yang, et~al. 2024.
\newblock Uncovering language disparity of chatgpt on retinal vascular disease classification: Cross-sectional study.
\newblock \emph{Journal of Medical Internet Research}, 26(1):e51926.

\bibitem[{Lyu et~al.(2023)Lyu, Tan, Zapadka, Ponnatapura, Niu, Myers, Wang, and Whitlow}]{llm-simplify-gpt4}
Qing Lyu, Josh Tan, Michael~E Zapadka, Janardhana Ponnatapura, Chuang Niu, Kyle~J Myers, Ge~Wang, and Christopher~T Whitlow. 2023.
\newblock Translating radiology reports into plain language using chatgpt and gpt-4 with prompt learning: results, limitations, and potential.
\newblock \emph{Visual Computing for Industry, Biomedicine, and Art}, 6(1):9.

\bibitem[{Nori et~al.(2023)Nori, King, McKinney, Carignan, and Horvitz}]{medical-gpt4-2023}
Harsha Nori, Nicholas King, Scott~Mayer McKinney, Dean Carignan, and Eric Horvitz. 2023.
\newblock Capabilities of gpt-4 on medical challenge problems.
\newblock \emph{arXiv preprint arXiv:2303.13375}.

\bibitem[{Papineni et~al.(2002)Papineni, Roukos, Ward, and Zhu}]{metrics-bleu}
Kishore Papineni, Salim Roukos, Todd Ward, and Wei-Jing Zhu. 2002.
\newblock Bleu: a method for automatic evaluation of machine translation.
\newblock In \emph{Proceedings of the 40th annual meeting of the Association for Computational Linguistics}, pages 311--318.

\bibitem[{Pino et~al.(2021)Pino, Parra, Besa, and Lagos}]{metrics-medical-report}
Pablo Pino, Denis Parra, Cecilia Besa, and Claudio Lagos. 2021.
\newblock Clinically correct report generation from chest x-rays using templates.
\newblock In \emph{Machine Learning in Medical Imaging: 12th International Workshop, MLMI 2021, Held in Conjunction with MICCAI 2021, Strasbourg, France, September 27, 2021, Proceedings 12}, pages 654--663. Springer.

\bibitem[{Radford et~al.(2019)Radford, Wu, Child, Luan, Amodei, Sutskever et~al.}]{baseline-gpt2}
Alec Radford, Jeffrey Wu, Rewon Child, David Luan, Dario Amodei, Ilya Sutskever, et~al. 2019.
\newblock Language models are unsupervised multitask learners.
\newblock \emph{OpenAI blog}, 1(8):9.

\bibitem[{Risch et~al.(2021)Risch, M{\"o}ller, Gutsch, and Pietsch}]{metrics-qa-2021}
Julian Risch, Timo M{\"o}ller, Julian Gutsch, and Malte Pietsch. 2021.
\newblock Semantic answer similarity for evaluating question answering models.
\newblock \emph{arXiv preprint arXiv:2108.06130}.

\bibitem[{Sarraju et~al.(2023)Sarraju, Bruemmer, Van~Iterson, Cho, Rodriguez, and Laffin}]{llm-consulation-cardio-jama}
Ashish Sarraju, Dennis Bruemmer, Erik Van~Iterson, Leslie Cho, Fatima Rodriguez, and Luke Laffin. 2023.
\newblock Appropriateness of cardiovascular disease prevention recommendations obtained from a popular online chat-based artificial intelligence model.
\newblock \emph{JAMA}, 329(10):842--844.

\bibitem[{Serban et~al.(2016)Serban, Sordoni, Bengio, Courville, and Pineau}]{baseline-hred}
Iulian Serban, Alessandro Sordoni, Yoshua Bengio, Aaron Courville, and Joelle Pineau. 2016.
\newblock Building end-to-end dialogue systems using generative hierarchical neural network models.
\newblock In \emph{Proceedings of the AAAI conference on artificial intelligence}, volume~30.

\bibitem[{Singhal et~al.(2023{\natexlab{a}})Singhal, Azizi, Tu, Mahdavi, Wei, Chung, Scales, Tanwani, Cole-Lewis, Pfohl et~al.}]{medical-medpalm}
Karan Singhal, Shekoofeh Azizi, Tao Tu, S~Sara Mahdavi, Jason Wei, Hyung~Won Chung, Nathan Scales, Ajay Tanwani, Heather Cole-Lewis, Stephen Pfohl, et~al. 2023{\natexlab{a}}.
\newblock Large language models encode clinical knowledge.
\newblock \emph{Nature}, 620(7972):172--180.

\bibitem[{Singhal et~al.(2023{\natexlab{b}})Singhal, Tu, Gottweis, Sayres, Wulczyn, Hou, Clark, Pfohl, Cole-Lewis, Neal et~al.}]{medical-medpalm-2}
Karan Singhal, Tao Tu, Juraj Gottweis, Rory Sayres, Ellery Wulczyn, Le~Hou, Kevin Clark, Stephen Pfohl, Heather Cole-Lewis, Darlene Neal, et~al. 2023{\natexlab{b}}.
\newblock Towards expert-level medical question answering with large language models.
\newblock \emph{arXiv preprint arXiv:2305.09617}.

\bibitem[{Sutskever et~al.(2014)Sutskever, Vinyals, and Le}]{baseline-seq2seq}
Ilya Sutskever, Oriol Vinyals, and Quoc~V Le. 2014.
\newblock Sequence to sequence learning with neural networks.
\newblock \emph{Advances in neural information processing systems}, 27.

\bibitem[{Thirunavukarasu et~al.(2023{\natexlab{a}})Thirunavukarasu, Ting, Elangovan, Gutierrez, Tan, and Ting}]{thirunavukarasu2023large}
Arun~James Thirunavukarasu, Darren Shu~Jeng Ting, Kabilan Elangovan, Laura Gutierrez, Ting~Fang Tan, and Daniel Shu~Wei Ting. 2023{\natexlab{a}}.
\newblock Large language models in medicine.
\newblock \emph{Nature medicine}, 29(8):1930--1940.

\bibitem[{Thirunavukarasu et~al.(2023{\natexlab{b}})Thirunavukarasu, Ting, Elangovan, Gutierrez, Tan, and Ting}]{llm-med-natmed}
Arun~James Thirunavukarasu, Darren Shu~Jeng Ting, Kabilan Elangovan, Laura Gutierrez, Ting~Fang Tan, and Daniel Shu~Wei Ting. 2023{\natexlab{b}}.
\newblock Large language models in medicine.
\newblock \emph{Nature medicine}, 29(8):1930--1940.

\bibitem[{Tu et~al.(2023)Tu, Azizi, Driess, Schaekermann, Amin, Chang, Carroll, Lau, Tanno, Ktena et~al.}]{tu2023towards}
Tao Tu, Shekoofeh Azizi, Danny Driess, Mike Schaekermann, Mohamed Amin, Pi-Chuan Chang, Andrew Carroll, Chuck Lau, Ryutaro Tanno, Ira Ktena, et~al. 2023.
\newblock Towards generalist biomedical ai.
\newblock \emph{arXiv preprint arXiv:2307.14334}.

\bibitem[{Wang et~al.(2023{\natexlab{a}})Wang, Zhao, Ouyang, Wang, and Shen}]{llm-chatcad}
Sheng Wang, Zihao Zhao, Xi~Ouyang, Qian Wang, and Dinggang Shen. 2023{\natexlab{a}}.
\newblock Chatcad: Interactive computer-aided diagnosis on medical image using large language models.
\newblock \emph{arXiv preprint arXiv:2302.07257}.

\bibitem[{Wang et~al.(2023{\natexlab{b}})Wang, Sanders, Liu, Seang, Tran, Atanasov, Qiu, Tang, Car, Wang et~al.}]{llm-lmic-lancet}
Xiaofei Wang, Hayley~M Sanders, Yuchen Liu, Kennarey Seang, Bach~Xuan Tran, Atanas~G Atanasov, Yue Qiu, Shenglan Tang, Josip Car, Ya~Xing Wang, et~al. 2023{\natexlab{b}}.
\newblock Chatgpt: promise and challenges for deployment in low-and middle-income countries.
\newblock \emph{The Lancet Regional Health--Western Pacific}, 41.

\bibitem[{Wei et~al.(2018)Wei, Liu, Peng, Tou, Chen, Huang, Wong, and Dai}]{dialogue-diagnosis-muzhi}
Zhongyu Wei, Qianlong Liu, Baolin Peng, Huaixiao Tou, Ting Chen, Xuan-Jing Huang, Kam-Fai Wong, and Xiang Dai. 2018.
\newblock Task-oriented dialogue system for automatic diagnosis.
\newblock In \emph{Proceedings of the 56th Annual Meeting of the Association for Computational Linguistics (Volume 2: Short Papers)}, pages 201--207.

\bibitem[{Wu et~al.(2023)Wu, Wu, Qiu, Li, Zheng, and Yang}]{cnmle-chatgpt-2023}
Jiageng Wu, Xian Wu, Zhaopeng Qiu, Minghui Li, Yefeng Zheng, and Jie Yang. 2023.
\newblock Qualifying chinese medical licensing examination with knowledge enhanced generative pre-training model.
\newblock \emph{arXiv preprint arXiv:2305.10163}.

\bibitem[{Xu et~al.(2023)Xu, Hou, Cheng, Wang, and Li}]{medical-dialogue-dfmed}
Kaishuai Xu, Wenjun Hou, Yi~Cheng, Jian Wang, and Wenjie Li. 2023.
\newblock \href {https://doi.org/10.18653/v1/2023.findings-acl.423} {Medical dialogue generation via dual flow modeling}.
\newblock In \emph{Findings of the Association for Computational Linguistics: ACL 2023}, pages 6771--6784, Toronto, Canada. Association for Computational Linguistics.

\bibitem[{Zhang et~al.(2019)Zhang, Kishore, Wu, Weinberger, and Artzi}]{metrics-bert}
Tianyi Zhang, Varsha Kishore, Felix Wu, Kilian~Q Weinberger, and Yoav Artzi. 2019.
\newblock Bertscore: Evaluating text generation with bert.
\newblock \emph{arXiv preprint arXiv:1904.09675}.

\bibitem[{Zheng et~al.(2023)Zheng, Chiang, Sheng, Zhuang, Wu, Zhuang, Lin, Li, Li, Xing et~al.}]{metric-llm-judge}
Lianmin Zheng, Wei-Lin Chiang, Ying Sheng, Siyuan Zhuang, Zhanghao Wu, Yonghao Zhuang, Zi~Lin, Zhuohan Li, Dacheng Li, Eric Xing, et~al. 2023.
\newblock Judging llm-as-a-judge with mt-bench and chatbot arena.
\newblock \emph{arXiv preprint arXiv:2306.05685}.

\end{thebibliography}
\bibliographystyle{acl_natbib}

% \appendix

% \section{Example Appendix}
% \label{sec:appendix}

% This is a section in the appendix.

\end{document}